\documentclass[preprint,3p,times]{elsarticle}

\usepackage{times,mathtools,amsfonts}
\usepackage{helvet,courier}
\usepackage{txfonts}
\usepackage{framed}
\usepackage{graphicx}
\usepackage{color}

\newtheorem{THEOREM}{Proposition}
\newenvironment{theorem}{\begin{THEOREM} }%
                        {\end{THEOREM}}

\newtheorem{LEMMA}[THEOREM]{Lemma}
                      {\end{LEMMA}}
\newtheorem{COROLLARY}[THEOREM]{Corollary}
                          {\end{COROLLARY}}
\newtheorem{PROPOSITION}[THEOREM]{Proposition}
                            {\end{PROPOSITION}}
\newtheorem{DEFINITION}[THEOREM]{Definition}
\newenvironment{definition}{\begin{DEFINITION}  \rm}%
                            {\end{DEFINITION}}
\newtheorem{CLAIM}[THEOREM]{Claim}
                            {\end{CLAIM}}
\newtheorem{EXAMPLE}[THEOREM]{Example}
\newenvironment{example}{\begin{EXAMPLE} \rm}%
                            {\end{EXAMPLE}}
\newtheorem{REMARK}[THEOREM]{Remark}
                            {\end{REMARK}}
							\newtheorem{NOTATION}[THEOREM]{Notation}
							                            {\end{NOTATION}}
                     {}

\DeclareMathAlphabet{\mathitbf}{OML}{cmm}{b}{it}

\newcommand{\sub}{_}
\def\su{^}

\newcommand{\real}{{\mathbb{R}}}

\newcommand{\nat}{{\mathbb{N}}}

\newcommand{\wmc}{\textsc{WMC}}

\newcommand{\amc}{\#}

\newcommand{\lt}{<}
\newcommand{\gt}{>}

\newcommand{\infinity}{\infty}
\newcommand{\A}{{\mathbb S}}
\renewcommand{\AA}{{\cal S}}

\newcommand{\C}{{\cal C}}
\newcommand{\D}{{\cal D}}

\renewcommand{\L}{{\cal L}}

\newcommand{\mm}{\models}
\newcommand{\M}{{\cal M}}

\renewcommand{\P}{{\cal P}}

\newcommand{\R}{{\cal R}}
\renewcommand{\S}{{\cal S}}
\newcommand{\T}{{\cal T}}

\newcommand{\X}{{\cal X}}

\newcommand{\Z}{{\cal Z}}

\newcommand{\set}[1]{\left\{ #1 \right\}}
\newcommand{\la}{\langle}
\newcommand{\ra}{\rangle}

\newcommand{\idenz}{{\mathbf 0}}
\newcommand{\ideno}{{\mathbf 1}}
\newcommand{\eg}{\emph{e.g.,}~}
\newcommand{\ie}{\emph{i.e.,}~}

\begin{document}
	 
	 \begin{frontmatter}

%
\title{Semiring Programming: \\
A Declarative Framework for Generalized Sum Product Problems
} 

		\cortext[cor]{Corresponding author. The author was supported by a Royal Society University Research Fellowship.}

		 \address[edinburgh]{School of Informatics, University of Edinburgh,
		Edinburgh, UK.}
		 \address[turing]{Alan Turing Institute,
		London, UK.}
		\address[leuven]{Department of Computer Science, KU Leuven, Belgium.}

		\author[edinburgh,turing]{Vaishak Belle\corref{cor}}
		\ead{vaishak@ed.ac.uk}
		
		\author[leuven]{Luc De Raedt}
		\ead{
luc.deraedt@cs.kuleuven.be}


\begin{abstract} 
	


%
	To solve hard problems, AI relies on a variety of disciplines such as logic,  probabilistic reasoning, machine learning and mathematical programming. Although it is widely accepted that solving real-world problems requires an integration amongst these, contemporary representation methodologies offer little support for this.

In an attempt to alleviate this situation, we introduce a new declarative programming framework that provides abstractions of well-known problems such as SAT, Bayesian inference, generative models, learning  and convex optimization. The semantics of programs is defined in terms of first-order logic structures with semiring labels, which allows us to freely combine and integrate problems from different AI disciplines and represent non-standard problems over unbounded domains. 

\end{abstract}


\end{frontmatter}

\section{Introduction} 
\label{sec:introduction}

AI applications, such as robotics and logistics, rely on a variety of disciplines such as logic, probabilistic reasoning, machine learning and mathematical programming. 
These applications are often described in a combination of natural and mathematical language, and need to be engineered for the individual application.  

Declarative formalisms and methods are ubiquitous in AI as they enable re-use and descriptive clarity.  
Initial approaches, such as that by   Kowalski~\cite{DBLP:conf/ifip/Kowalski74},  were
rooted in logic, but they have eventually engendered an impressive family of languages. In knowledge representation and constraint programming, for example, languages such ASP \cite{brewka2011answer} and  Essence \cite{DBLP:journals/constraints/FrischHJHM08} are prominent, which use SAT, SMT  and MIP technology \cite{BSST09HBSAT,Sebastiani:2015:OMT:2737801.2699915}. In machine learning and probabilistic reasoning, statistical relational learning systems
and probabilistic programming languages such as Markov Logic \cite{richardson2006markov}, Church \cite{conf/uai/GoodmanMRBT08} and Problog \cite{DBLP:conf/ijcai/RaedtKT07} are increasingly used to codify intricate inference and learning tasks. In mathematical programming and optimization, disciplined programming \cite{gb08} and AMPL \cite{opac-b1123349} have been developed. Finally, 
the  DARPA project \emph{Probabilistic Programming for Advancing Machine Learning}  is motivated in the same declarative spirit.\footnote{\texttt{http://www.darpa.mil/program/probabilistic-programming-for-advancing-machine-learning}}
Across  disciplines in AI, it has  become increasingly clear that taming the model building process, admitting reusable descriptions in expressive languages, 
and providing general but powerful inference engines is essential. 

Be that as it may, it is widely accepted that solving real-world AI problems requires an integration of different disciplines. Consider, for example, that a robot may decide its course of action using a SAT-based planner, learn about the world using Kalman filters, and grasp objects using  geometric optimization technology. But contemporary declarative frameworks offer little support for such universality: knowledge representation and constraint formalisms mostly focus on model generation for discrete problems, probabilistic programming languages do not handle linear and arithmetic constraints, and finally, optimization frameworks work with linear algebra and algebraic constraints to specify the problem and thus are quite different from the high-level descriptions used in the other disciplines and do not support probabilistic or logical reasoning.

Of course, hybrid approaches that treat different computations as independent but communicating processes is an option to address this challenge, but these integrations may not be transparent. So what is lacking here is a universal modeling  framework that allows us to declaratively specify problems involving \emph{logic} and \emph{constraints}, \emph{mathematical programs}, as well as \emph{discrete
and continuous probability distributions} in a simple, uniform, modular and transparent manner. Such a framework, together with a generic inference mechanism, would greatly simplify the development and understanding of AI systems with integrated capabilities, and would tame the model building process. 


There has been recent  progress on this front. A key observation made in \cite{AMC} is that reasoning about possible worlds is fundamental to many tasks in computer science, including dynamic programming \cite{eisner-filardo-2011}, constraint programming \cite{Bistarelli:2004:SSC:993462}, database theory \cite{green2007provenance}, probabilistic inference \cite{DBLP:journals/jair/BacchusDP09}, probabilistic logic programming \cite{DBLP:conf/ijcai/RaedtKT07}, and network analysis \cite{baras2010path}. In fact, these tasks essentially invoke a version of the sum product problem \cite{DBLP:journals/jair/BacchusDP09},  but differ in the exact operations carried out over the possible worlds, which can then be recast in the very same way via semirings. The resulting framework, referred to as \textit{algebraic model counting} (AMC)  \cite{AMC}, shows  that  the computation underlying all these tasks can be defined over a certain class of arithmetic circuits, which then implies that they can be solved via a  single algorithm that obtains local solutions and composes them to yield a global solution. The main limitation of the AMC proposal is that the underlying language is propositional logic, and so the semantics is that of classical logic and computations are essentially defined over discrete spaces. 

In this paper, we propose a new declarative framework called \emph{semiring programming} (SP) that  attempts to generalize AMC. Our main thesis is to still formulate computations as the sum product problem, but we will rigorously define a semantics that  not only defines AMC in \emph{unbounded} domains, but also  \emph{non-standard} (e.g., non-monotonic) ones. To put the proposal in perspective, consider that Eugene Freuder \cite{freuder} famously quipped: ``constraint programming represents one of the closest approaches computer science has yet made to the Holy Grail of programming: the user states the problem, the computer solves it." The underlying idea was summarized in the slogan: \[
	\textrm{\bf (constraint) program = model + solver}
\]   The vision of SP builds upon this equation in that: \[
	\textrm{\bf (semiring) program = logical theory + semiring + solver}
\] Thus a model is expressed in a logical system in tandem with a semiring and a weight function. As a framework, SP is set up to allow the modeler to freely choose the logical theory (syntax and semantics) and so everything from non-classical logical consequence to real-arithmetic is fair game. Together with a semiring and these weights, a program computes the \emph{count}. 
Our task will be to show that the usual suspects from AI disciplines, such as SAT, CSP, Bayesian inference, and convex optimization can be: (a) expressed as a program, and (b) count is a solution to the problem, that is, the 
count can be a \{\emph{yes}, \emph{no}\} answer in SAT, a probability in Bayesian inference, or a bound in convex optimization. In other words, the count is shown to \emph{semantically  abstract} challenging AI tasks. We will also demonstrate a variety of more complex problems, such as matrix factorization, and one on compositionality. 

Informally, from an expressiveness viewpoint, SP is designed to be: 

\begin{itemize}
	\item {\em universal}, in  that it 
	can represent classical problems from disciplines ranging from logic to mathematical programming, and 
	it inherits the strengths of both camps;  
	\item {\em declarative}, which means that domain knowledge can be expressed in a \emph{program} in a natural, human-readable way, and that it is possible to easily cope with changes in requirements and information in a principled manner; 
	\item {\em generic}, in that it permits instantiations to a particular language -- including real arithmetic (as needed in machine learning),  
quantification and interpreted symbols (as in satisfiability modulo theory, or  SMT), and non-classical  consequence (as in answer set programming, or ASP) -- since these often correspond closely to the kind of  problems they attempt to formalize and solve; 
       \item {\em solver-independent},  in  that inference assumes the role of  algorithms, and the formulation of the inference problems is separate from the solution strategies;
	\item {\em model-theoretic}, in that the semantics of programs is defined using first-order structures, in service of providing meaning to  classical model generation problems ranging from SAT to convex optimization. 
\end{itemize} 
To reiterate, the aim is to synthesize problems and techniques across important disciplines in AI, towards a modeling framework that allows one to freely combine 
and integrate a wide range of  specifications. To our knowledge, a number of prior proposals have made promising progress towards these goals, 
but have fallen short in certain critical features; see the penultimate section for discussions. 

We reiterate that the thrust of the proposal is to rigorously generalize standard AMC, so as to {semantically  abstract} problems arising from model generation in propositional satisfiability, first-order declarative programming, machine learning, data mining, constraint satisfaction, and optimization. In attempting such a generalization, of course, we will have to give up the existing investigations on the use of propositional arithmetic circuits, including the simple evaluation scheme on local solutions informing global ones \cite{AMC}. That is not surprising given the infinitary nature of the framework, but we hope the framework will provide the necessary  foundations for investigating a general solver scheme that works on both finite and infinite domains. Moreover, we illustrate the framework using a programming language inspired by SMT syntax, but this is meant to be a prototypical starting point for a full fledged programming interface to SP. As can be inferred from above, we would consider the following three desiderata to be essential in the design of such an interface: \begin{enumerate}
	\item The logical language should be expressive, supporting a rich ontology of types including lists and graphs.
	\item The language should have a semantics defined by a measure on a set of worlds respecting semiring operators. 
	
	\item The language should provide abstractions for specifying the solutions of (arbitrary) problems involving \emph{logical reasoning}, \emph{discrete-continuous probability distributions} and  \emph{discrete}-\emph{continuous}  \emph{optimization formulations} in a simple, uniform, modular and transparent manner. 
\end{enumerate} 
The formulation in the sequel will provide insights on how such a language could be obtained. 

This paper is organised as follows. In Section \ref{sub:logical_preliminaries}, we briefly review logic, weighted model counting and semirings; in Section \ref{sec:algebraic_model_counting};  we introduce the semiring programming framework and illustrate it on a number of examples; in Section \ref{sec:combining_theories}, we discuss how different semirings can be combined;  in Section \ref{sec:solver_strategy} different strategies for solvers are introduced; and finally, in Sections \ref{sec:related_approaches} and 
 \ref{sec:discussion} we discuss related work and conclude.

\section{Environments} 
\label{sub:logical_preliminaries}




\renewcommand{\models}{\rhd}

\subsection{Logical Setup} 
\label{sub:logical_setup}

The framework is developed in a general way, agnostic about the meaning of sentences. We adopt (and assume familiarity with)  terminology from predicate logic \cite{enderton1972mathematical}.





%
%
%
%

\newcommand{\sig}{\textit{vocab}}
\newcommand{\dom}{{\it dom}}

\newcommand{\lits}{\textit{lits}}

\begin{definition} A \emph{theory} \( \T \) is a triple $(\L, \M, \models)$, where \( \L \) is a set of sentences called the \emph{language} of the theory, \( \M \) a set called the \emph{models} of the theory, and \( \mm \) a subset of \( \M\times \L \) called the \emph{satisfiability relation}. 
	
\end{definition} 

The set \( \L \) is implicitly assumed to be defined over a vocabulary \( \sig(\L) \)  of relation and function symbols, each with an associated arity. Constant symbols are 0-ary function symbols. Every \( \L \)-model \( M\in\M \) is a tuple containing a universe \( \dom(M) \), and a relation (function) for each relation (function) symbol of \( \sig(\L) \). For relation (constant) symbol \( p \), the relation (universe element) corresponding to \( p \) in a model \( M \) is denoted \( p\su M. \) 
%
%
{For \( \phi \in\L \) and \( M\in \M, \) we write \( M \mm \phi \) to say that \( M \)  \emph{satisfies} \( \phi. \)} We say a formula $\phi$ is \textit{valid}  iff $\phi$ is satisfied at every model, which is then written as $\models \phi.$ 
We let \( \M{(\phi)} \) denote \(\set{ M \in \M \mid M \mm \phi }\). 
Finally, the set \( \lits(\L) \) denotes the literals in \( \L, \) and  we write  \( l\in M \) to refer to the \( \L \)-literals that are satisfied at \( M \). 



This formulation is henceforth used to instantiate a particular logical system, such as fragments/extensions of first- order logic, as well as non-classical logical consequence.

\begin{example}\label{ex:prop theory} Suppose \( \T = (\L,\M, \mm) \) is as follows: \( \L \) is obtained using a set of 0-ary predicates \( \P \) and Boolean connectives, \( \M \) are \( \set{0,1} \) assignments to the elements of \( \P \), and \( \mm \) denotes satisfaction in propositional logic via the usual inductive rules. Then, we obtain a \emph{propositional theory}.	
\end{example}

\begin{example}\label{ex:minimal models} Define a theory \( (\L, \M, \mm) \) where \( \L \) is the positive Horn fragment from a finite vocabulary, \( \M \) the set of propositional interpretations, and  for \( M\in \M, \phi\in \L \), define \( M\mm \phi \) iff \( M \models\sub{\rm PL} \phi \) (\ie satisfaction in propositional logic) and for all \( M' \subsetneq M,  \) it is not the case that \( M' \models\sub{\rm PL} \phi. \) This theory can be used to reason about  the \emph{minimal models} of a formula. 
	
\end{example}

\begin{example} Define a theory \( (\L, \M, \mm) \) where \( \L \) is a first-order language involving 0-ary functions \( \set{c,\ldots,d} \), inequalities \( \leq, \geq, \lt, \gt, =, \neq. \) Let \( \M \) be the set of mappings from \( \set{c,\ldots,d} \) to \( \real \). We define \( M \models \phi \)  for \( M\in \M \) and \( \phi \in \L \) as in first-order logic, assuming in particular that  
$=,\lt,\gt,0,1,+,\times,/,-,$  exponentiation and logarithms have their usual interpretations  \cite{BSST09HBSAT}. 
(That is,``$1+0=1$"  is true in all models, as is ``$x>y \equiv \neg(y>x)$," and so on \cite{BELLE2018189}.) 
	 This theory can be used to reason about linear arithmetic (i.e., allowing formulas such as \( c + d \leq 5  \) and $d \leq e$, where $c,d,e$ are integers or reals) and non-linear arithmetic (i.e., allowing formulas such as  \( c + d\su 2 \geq e \)). 
	 
	
\end{example}

%


\subsection{Weighted Model Counting} %
\label{sub:weighted_model_counting}

Semiring programming   draws from the conceptual simplicity of weighted model counting (WMC), which we briefly recap here. WMC is an extension of \#SAT, where one simply counts the number of models of a propositional  formula \cite{modelcountingchapter}. In WMC, one accords a weight to a model in terms of the literals true at the model, and computes the sum of the weights of all models.

\begin{definition}\label{defn wmc} 
Suppose \( \phi \) is a formula from  a  propositional language \( \L \) with a finite vocabulary, and suppose \( \M \) is the set of \( \L \)-models. Suppose \( w\colon  \emph{lits}(\L)  \rightarrow \real\su {\geq 0} \)  is a weight function. Then:  \[
	\wmc(\phi,w) = \sum\sub {M\models \phi} \prod\sub {l\in M} w(l)
\] 
is called the \emph{weighted model count} (WMC) of \( \phi \). 
\end{definition}

Here, {in the context of \( \L \)-models \( \M, \)
we simply write \( t = \sum\sub {M\models \phi} u \) to mean \( t = \sum\sub {\set{M\in\M\mid M\models\phi}} u \).}

The formulation elegantly decouples the logical sentence from the weight function. In this sense, it  
is clearly agnostic about how weights are specified in the modeling language, and thus, has emerged as an assembly language for Bayesian networks~\cite{DBLP:journals/ai/ChaviraD08},   and  probabilistic  programs \cite{DBLP:conf/uai/FierensBTGR11}, among others.

\subsection{Commutative Semirings} %
\label{sub:commutative_semirings}

Our programming model is based on algebraic structures called \emph{semirings} \cite{MR98m:68152}; the essentials are as follows: 

\begin{definition} A (commutative) semiring \( \AA \) is a structure \( (\A, \oplus, \otimes, \idenz, \ideno) \) where \( \A \) is a set called the \emph{elements} of the semiring,  \(  {\oplus} \) and \( \otimes \) are associative and commutative,  \( \idenz \) is the identity for \( \oplus, \) and \( \ideno \) is the identity for \( \otimes. \) 
	
\end{definition}

Abusing notation, when the multiplication  operator is not used, we simply refer to the triple \( (\A, \oplus, \idenz) \) as a semiring. 

\begin{example} The structure \( (\nat, +, \times, 0, 1) \) is a commutative semiring in that for every \( a, b \in \nat, \) \( a+0 = a, \) \( a\times 1 = a \), \( a+b = b+a,  \) and so on. 
\end{example}

\section{Semiring Programming} %
\label{sec:algebraic_model_counting}

In essence, the semiring programming scheme is as follows: \begin{itemize}
	\item \textbf{Input}: a theory \( \T = (\L, \M, \mm) \), a sentence \( \phi \in \L,  \) a commutative semiring \( \AA \), and a weight function \( w. \)   
	\item \textbf{Output}: the \emph{count}, denoted $\#(\phi,w)$. 
\end{itemize}
The scope of these programs is broad, and so we will need different kinds of generality. Roughly, the distinction boils down to: (a) whether the set of models for a formula is \emph{finite} or \emph{infinite}; (b) whether the weight function can be \emph{factorized} over the literals or not (in which case the weight function directly labels the models of a theory); and (c) whether \( \models \) is defined in a classical (monotonic) manner or not. 
 The thrust of this section is: (i) to show how these distinctions subsume important model generation notions in the literature, and (ii) providing rigorous definitions for the count operator. In terms of organization, we begin with the finite case, before turning to the infinite ones. We present an early preview of some of the models considered  in Figure \ref{fig:scope}.

\begin{center}
	\begin{figure}[t]
			\begin{tabular}{c|c|c|c}
			              & factorized & non-factorized & non-classical \\
			\hline
			finite & WMC &  logical-integer programming & ASP \\
			infinite            & polyhedron volume & convex optimization & first-order ASP  \\
			\end{tabular}
			\caption{scope of programs}\label{fig:scope}
	\end{figure}
\end{center}

\begin{figure}
	\begin{framed}\small\tt
		(set-logic PL) \\\
		(set-algebra [NAT,max,*,0,1]) \\
		(declare-predicate p ())\\
		(declare-predicate q ())\\
		F = (p or q)\\
		(declare-weight (p 1))\\
		(declare-weight ((neg p) 2)) 		\\
		(declare-weight (q 3))\\
		(declare-weight ((neg q) 4))\\
		(count F)
	\end{framed} \caption{most probable explanation}\label{fig:sat}
\end{figure}

\subsection{Finite} %
\label{sub:finitely_countable}

Here, we generalize the \wmc\ formulation to {semiring} labels, but also go beyond classical propositional logic.

\begin{definition} We say the theory \( \T = (\L,\M,\mm) \) is \emph{finite} if for every \( \phi \in \L, \) \( \set{M\in \M \mid M \mm \phi} \) is finite.	
\end{definition}
So, a propositional language with a finite vocabulary is a finite theory, regardless of (say) standard or minimal models. Similarly, a first-order language with a finite Herbrand base is also a  finite theory. %

\begin{definition}\label{defn finite amc} Suppose  \( \T = (\L, \M, \mm) \) is  a  finite theory.  Suppose \( \AA = (\A, \oplus, \otimes, \idenz, \ideno) \) is a  {commutative semiring}. Suppose \( w: \M \rightarrow \A. \) For any \( \phi\in \L, \) we define: \[
	\amc(\phi, w) = \bigoplus\sub {M\mm \phi} w(M)
\]
If \( w :\lits(\L) \rightarrow \A \), then the problem is   \emph{factorized}, where:\[ w(M)  =  \bigotimes\sub {l\in M} w(l). \]
\end{definition}
Essentially, as in \wmc, we sum over models and take products of the weights on literals but w.r.t. a particular semiring. Needless to say, we immediately subsume the framework of algebraic model counting (AMC) \cite{AMC}:

\begin{theorem} Suppose  \( \T = (\L, \M, \mm) \) is  a  finite propositional theory,  
	\( \AA \) a commutative semiring, and \( w \) a factorized weight function. Suppose \( \mm \) is the  standard satisfaction relation in propositional logic. Then SP is equivalent to AMC, that is, the computation of every SP instance can be defined as an AMC task and vice versa. 
\end{theorem}

Let us consider a few examples. 

\begin{example}\label{ex:sat as amc}  We demonstrate SAT and \#SAT. 
	Consider a propositional theory \( \T = (\L, \M, \models) \) where \( \sig(\L) = \set{p,q}, \) and suppose \( \phi = (p\lor q). \) Letting \( \M \) be standard \( \L \)-models, clearly \( |\M| = 4 \) and \( |\set{M \in \M\mid M \models \phi}| = 3 \). Suppose for every \( M\in \M,  \) \( w \) is function such that \( w(M) = 1. \) For the semiring \( (\set{0,1}, \lor, 0) \): \[ \#(\phi, w) = \bigvee\sub {M\models \phi} w(M) = 1 \lor 1 \lor 1 = 1. \] 
	
	Consider the semiring \( (\nat, +, 0) \) instead. Then  \[
		\#(\phi, w) = \sum\sub{M\models \phi} w(M) = 1 + 1 +1 =  3.
	\]
\end{example}

As with Definition \ref{defn wmc}, the framework is  agnostic about the modeling language. But for presentation purposes, programs are sometimes described using a notation inspired by the SMT-LIB standard \cite{BarFT-RR-15}.
	
\begin{example}	We demonstrate  MPE (most probable explanation) and WMC. Consider the theory,  semiring  and   weight function \( w \) from Figure \ref{fig:sat}, which specifies, for example, a vocabulary of two propositions \( p \) and \( q, \) 
	\( w(p) = 1 \)  and \( w(\neg p) =2. \) In accordance with that  semiring,  the weight of a model, say \( \set{p,\neg q} \),  of the formula \( \texttt F \) is \(  1\times 4 = 4. \)
	Thus, for the semiring \( (\nat, \max, \times, 0, 1) \), we have: \[
	\#({\texttt F},w) = \max \set{6,3,4} = 6
\]
which finds the most probable assignment.  
Consider the semiring \( (\nat,+,\times,0,1) \) instead. Then: \[
		\#({\texttt F},w) = 6 + 3 + 4 = 13
	\]
which gives us the weighted model count. 	

\end{example}

\begin{example} Extending the discussion in \cite{DBLP:journals/dam/HookerO99}, we consider a class of mathematical programs where linear constraints and propositional formulas can be combined freely. See Figure \ref{fig:logic mathemtical programming} for an example with non-linear objectives. 
	Formally, quantifier-free linear integer arithmetic and propositional logic are specified as the underlying logical systems, and the domains of constants are typed. The program declares formulas $\texttt F$, $\texttt G$, and \( {\texttt H}\).
	 
	 The counting task is non-factorized, and our convention for assigning weights to models is by letting the \[\text{\tt declare-weight} \]  directive also take arbitrary formulas as arguments. Of course, {\tt TRUE} holds in every model, and so, the weight of every model is determined by the evaluation of \( \texttt {x1}* \texttt{x2} \) at the model, that is, for any \( M, \) its weight is \( {\texttt *}\su M ({\texttt {x1}}\su M, {\texttt {x2}}\su M)  \). For example, a model that assigns {\tt 1} to {\tt x1} and {\tt 1} to {\tt x2} is accorded the weight \(\texttt 1 * \texttt 1.  \)  Computing the count over \( (\nat,\max, 0) \) then yields a model of \( \texttt H \) with the highest value for \( \texttt {x1}*\texttt{x2} \).

To see this program in action, consider that every model of \( \texttt H \) must satisfy \(\texttt {p2}, \) and so must admit \( \texttt {3*x1} <= \texttt{4} \) and \(\texttt {2*x2} <= \texttt{5}. \) Since \( \texttt {\{x1,x2\}} \) can only take values   \texttt{\{1, $\ldots$,10\}}, given the constraints, the desired model must assign \( \texttt {x1} \) and \( \texttt {x2} \) to \( \texttt 1 \) and \( \texttt 2 \) respectively. Then, its weight is \( \texttt {1*2}. \)

\end{example}

\begin{figure}
	\begin{framed}\small\tt
		(set-logic QF\_LIA;PL) \\\
		(set-algebra [NAT,max,0]) \\
		(set-type INT=\{1,...,10\}) \\
		(declare-function x1 () INT) \\
		(declare-function x2 () INT) \\	(declare-predicate p1 ())\\
		(declare-predicate p2 ())\\
		F = ((p1 or p2) => 3*x1 <= 4) \\
		G = (p2 => (2*x2 <= 5)) \\
		H = ((F and G) and p2) \\
		(declare-weight  TRUE x1*x2)\\
		(count H) 		
	\end{framed} \caption{logical-integer programming}\label{fig:logic mathemtical programming}
\end{figure}

 Encoding finite domain constraint satisfaction problems as propositional satisfiability is well-known. The benefit, then, of appealing to our framework is the ability to easily formulate counting instances:

	\begin{theorem}\label{thm:csp} 
	 Suppose \( Q \) is a CSP over variables \( \X  \), domains $\D$ and constraints \( \C \). There is a  first-order theory \(  (\L,\M,\models) \), a \( \L \)-sentence \( \phi \) and a weight function \( w \) such that \( \amc(\phi,w) = 1 \) over \( (\set{0,1},\lor,0) \) iff \( Q \) has a solution. Furthermore, \( Q \) has \( n \) solutions iff \( \amc(\phi,w) = n \) over \( (\nat,+,0). \)	
		\end{theorem}			
	Constraints are  Boolean-valued functions  \cite{freuder2006constraint}, and so  constraints over \( \X \) can be  encoded as  \( \L \)-sentences, as in the example below: 
	

\begin{example} See Figure \ref{fig:graph coloring} for a counting instance of graph coloring: \( \texttt {edge(x,y)} \) determines there is an edge between {\tt x} and {\tt y}, {\tt node(x)} says that {\tt x} is a node, and {\tt color(x,y)} says that  node \( \texttt x \) is assigned the color \( \texttt y. \) The actual graph is provided using the formula \( \texttt {DATA}, \) which declares a fully connected 3-node graph. Also, {\tt CONS} is a conjunction of the usual  coloring constraints, \eg an edge between nodes \( \texttt x \) and \( \texttt y \) means that they cannot be assigned the same color.

	Let \( \M \) be a set of first-order structures for the vocabulary \( \set{\texttt{edge, node, color}} \), respecting types from Figure \ref{fig:graph coloring}. The interpretation of \{{\tt edge}, {\tt node}\} is assumed to be the same for all the models in \( \M \) and is as given by \( \texttt {DATA}. \) 
	Basically, then, the models differ in their interpretation of \( \texttt {color} \). One model of \(\phi = \) {\tt (DATA and CONS)}, for example, is \( \set{\texttt{color(1,r), color(2,g), color(3,b)}} \). The weights of all models is 1, and so, for \( (\nat, +, 0) \) we get:\footnote{In an analogous fashion, soft and weighted CSPs can be expressed  using semirings \cite{DBLP:journals/constraints/BistarelliMRSVF99}.}  \[ \#(\phi,w) = 6. \]

\end{example}

To summarize, the following result is easily shown for semiring programs:\footnote{
We remark that  although the count operator in itself does not provide the variable assignments for  SAT, MPE and CSP, we assume this can be retrieved from the satisfying interpretations. 
}  

\begin{theorem} Suppose \( \theta \in \{\rm SAT, \#SAT, WMC, MPE, CSP, \#CSP\} \). Suppose \( \theta\su \circ \) is a solution to \( \theta, \) in that 
	\( \theta\su \circ \in \set{0,1} \) for {\rm SAT} and {\rm CSP}, and \( \theta\su \circ \in \real \) for the rest. Then for any \( \theta \),  there is a \( \T = (\L, \M, \models) \), \( \AA, \) \( w \) and \( \phi\in \L \) such that \( \#(\phi,w) = \theta\su \circ. \) 
	
\end{theorem}

\begin{figure}
	\begin{framed}\small\tt
		(set-logic FOL) \\\
		(set-algebra [NAT,+,0]) \\
		(set-type COLOR=\{r,b,g\}) \\
		(set-type NODE=\{1,2,3\}) \\
		(declare-predicate node (NODE)) \\
		(declare-predicate edge (NODE,NODE))\\
 		(declare-predicate color (NODE,COLOR)) \\
	 N = (node(1) and node(2) and node(3)) \\
	 E = (edge(1,2) and edge(2,3) and edge(3,1)) \\
	 DATA = (N and E)\\
	 CONS = /* coloring constraints (omitted) */ \\
	 (declare-weight TRUE 1) \\		
	 (count (DATA and CONS))
	\end{framed} \caption{counting graph coloring instances}\label{fig:graph coloring}
\end{figure}

\subsection{Non-standard} 
\label{sub:non_standard_models}

A particular advantage of defining a logical theory in the way we did is that the framework is immediately applicable to model-level operations with non-standard semantics. We give a notable example that is simple to capture but which 
to the best of our knowledge has not been previously  formulated in a semiring  framework like ours. 

\begin{definition} A \emph{stable model environment} is defined as follows. 
	Let \( \T = (\L, \M, \models) \) be a logical theory, where \( \L \) is defined over a set of propositions \( \P \) and \( \M \) is the set of mappings from \( \P \) to \( \set{0,1} \). Let us split \( \P \)  into two disjoint sets of variables founded variables \( \P\sub f \) and standard variables \( \P\sub s \). An answer set program \( \delta \) is a tuple $(\P, \R, \C)$ where $\R$ is a set of rules of form: $a \leftarrow b\sub 1 \land \ldots \land b\sub n \land \neg c\sub 1 \land \ldots 
	\land \neg c\sub m$ such that $a \in  \P\sub f$, the body variables $\set{b\sub 1,\ldots,c\sub m} \subseteq  \P$, and $\C$ is a set of constraints over the propositions (specified as rules with an empty head). A rule is positive if its body only contains positive founded literals. The least assignment of a set of positive rules $\R$, written $L(\R)$ is the one that that satisfies all the rules and contains the least number of positive literals. Given an assignment \( M \in \M \) and a program \( \delta \), the reduct of \( M \) wrt \( \delta \), written, \( \delta\su M \) is a set of positive rules that is obtained as follows: for every rule \( r \), if any $c\sub i \in M $, or $\neg b\sub j \in M$ for any standard positive literal, then $r$ is discarded, otherwise, all negative literals and standard variables are removed from $r$ and it is included in the reduct. An assignment \( M \in \M \) is a stable model of a program \( \delta \) iff it satisfies all its constraints and \( M \)  \emph{agrees} with \( L(\delta\su M) \) wrt \( \P\sub f \). We say two assignments \( M \) and \( M' \) agree wrt \( \P' \) iff  the set of positive literals (restricted to \( \P' \)) in \( M \) is identical to the set in \( M' \), and the set of negative literals (restricted to \( \P' \)) in \( M \) is identical to the set in \( M'. \) For a program \( \delta, \) we let \( M \models \phi \) iff \( M \) is a stable model of \( \delta. \)
	
\end{definition}

This is basically an adaptation of the answer set programming by SAT formulation in \cite{aziz2015stable}. By way of the semirings considered in Example \ref{ex:sat as amc}, it immediately follows that: 

\begin{theorem} Suppose  \( \T = (\L, \M, \models) \) is a stable model environment, \( \delta\in \L \) is a program (as defined above). Suppose  \( \AA \) is the semiring $(\set{0, 1} , \lor, 0)$ and \( w(M) = 1 \) for every \( M\in \M. \) Then \( \#(\delta,w) \) tells us whether there is a stable model. For the semiring \( (\nat,+,0) \), \( \#(\delta,w) \) yields the number of stable models. 	
\end{theorem}

It is important to note that the formulation only semantically characterizes  the task of finding stable models as well as stable model counting. Algorithmic solutions for the task may very well involve other ideas \cite{aziz2015stable}. 

%
%

%
%


\subsection{Infinite: Non-factorized} %
\label{sub:a_simplification}

Defining measures \cite{halmos-measure} on the predicate calculus is central to logical characterizations of probability theory  \cite{halpern1990analysis,RupakSMT}. We adapt this notion for semirings to introduce a general form of counting. For technical reasons, we assume that the universe of the semirings is \( \real \). (If required, the range of the weight function can always be restricted to any subset of \( \real. \))

\begin{definition}\label{defn amc measure simple} Let \( \AA = (\real, \oplus, \idenz) \) be any semiring and \( \T = (\L, \M, \mm) \) a theory. Let \( \Sigma \) be a \( \sigma \)-algebra over \( \M \) in that \( (\M, \Sigma) \) is a \( \sigma \)-finite measurable space wrt the measure \( \mu\colon \Sigma \rightarrow \real \) respecting \( \oplus. \) That is, for all \( E \in \Sigma,  \) \( \mu(E) \geq 0, \) \( \mu(\set{}) = 0 \) and \( \mu \) is closed under complement and countable unions: for all pairwise disjoint countable sets \( E\sub 1, \ldots \in \Sigma, \) we have \[ \mu( \bigcup\sub {j=1}\su {\infinity} E\sub j) = \bigoplus\sub {j=1}\su \infinity \mu(E\sub j). \] Moreover, because the spaces are \( \sigma \)-finite, \( \M \) is the countable union of measurable sets with finite measure. 
Then for any \( \phi \in \L,  \)  \( \amc(\phi, \mu) = \mu(\M(\phi)). \)
\end{definition}

\begin{example}\label{ex:convex} We demonstrate that convex optimization can be cast as a semiring programming problem. 
		 Suppose \( \T = (\L, \M, \models) \) is a first-order theory only containing constants \(\X=  \set{x\sub 1, \ldots, x\sub k} \) with domains \( D\sub i = \real. \) Suppose \( \phi(x\sub 1, \ldots, x\sub k) \in \L \) is a conjunction of formulas of the form \( c\sub 1 x\sub 1 + \ldots + c\sub k x\sub k \leq d \), for real numbers \( c\sub 1, \ldots, c\sub k, d \), and thus describing a polyhedron. In other words, for every \( M\in\M, \) \( \dom(M) =\real, \) and so, \( M \) is a real-valued assignment to \( \X. \) In particular, if \( M\models \phi, \) then \( x\sub 1 \su M, \ldots, x\sub k \su M \) is a point inside the polyhedron \( \phi \). Let \( \Sigma \) be a \( \sigma \)-algebra over \( \M \) and so every \( E\in\Sigma \) is a measurable set of points.

		 Suppose \( f(x\sub 1, \ldots, x\sub k)\colon \real\su k \rightarrow \real \) is a convex function that we are to minimize. Consider the semiring \( (\real,\inf, 0) \) and a measure \( \mu \) such that for any \( E\in\Sigma \): \[
	\mu(E) = \inf \set{ f(x\sub 1 \su M, \ldots, x\sub k\su M) \mid M\in E}
\]
which finds the infimum of the \( f \)-values across the assignments in \( E. \) Then, \( \mu(\M(\phi)) = \amc(\phi, \mu) \) gives the minimum of the convex function in the feasible region determined by \( \phi. \) 

Suppose \( f \) is a concave function  that is to be maximized. We would then use \( (\real,\sup, 0) \) instead, which finds the supremum of the \( f \)-values across assignments in \( E\in \Sigma \). %

\end{example}

More generally, the same construction is easily shown to be applicable for other families of mathematical programming (\eg non-linear optimization) as follows: 

\begin{theorem}\label{thm math programming} Suppose \( \set{x\sub 1, \ldots, x\sub k} \) is a set of real-valued variables. Suppose \( P \subset \real\su k \) is the feasible region of an optimization problem of the form \( g\sub i (x\sub 1, \ldots, x\sub k) \circ d\sub i \) for \( i\in \nat, \) where \( \circ\in \set{\leq, \lt, \geq, \gt} \) and \( g\sub i\colon\real\su k\rightarrow \real \). Suppose \( f\colon\real\su k\rightarrow \real \) is a function to be maximized (minimized). Then there is a \( \T = (\L, \M, \models), \AA, \mu \) and \( \phi\in \L \) such that \( \#(\phi,\mu) \) is the maximum (minimum) value for \( f \) in the feasible region \( P. \) 
	
\end{theorem}

\begin{example}\label{ex matrix factorization} For a non-trivial  example, consider the problem of matrix factorization,   a  fundamental concern in information
retrieval and computer vision  \cite{Ding:2010:CSM:1687044.1687110}.  Given a matrix \( I \in \real\su {p \times n} \), we are to compute matrices \( L \in \real\su{p\times k} \) and \( R\in \real\su {n\times k} \), such that \( e = || I - LR\su T || \) is minimized. Here, \( || \cdot || \) denotes the Frobenius Norm. Using  real arithmetic, 
we provide a formulation in Figure \ref{fig:matrix factorization}. (Free variables are assumed to be implicitly quantified from the outside.)

Let \( \T = (\L, \M, \models) \) be the theory of real arithmetic, where \( \L \) includes the following  function symbols: \[ \{\texttt {input, left, right, app, err}\}. \] Here, \( \texttt {input(x,y)}  \) is a real-valued function such that  \( {\texttt x} \in \set{1, \ldots, p} \) and \( {\texttt y} \in \set{1, \ldots, n} \) in that \( \texttt {input(m,n)} \) is the entry at the \( m\su{\rm th} \) row and the \( n\su {\rm th} \) column of the matrix \( I \); these entries are specified by \( \texttt {DATA}. \) Letting \( \phi =  \) ({\texttt {DATA and F and G}}), the set \( \set{M\in\M \mid M \models \phi} \) are those \( \L \)-models whose interpretation of \( \texttt {input} \) is fixed by \( \texttt {DATA}. \) Basically, these models vary in their interpretations of \( \texttt {\{left, right\}} \), which determines their interpretations for \( \texttt {app} \) and \( \texttt {err} \). Here, {\tt app}   computes the product of the matrices {\tt left} and \( \texttt {right}, \) and {\texttt {err}} computes the Frobenius Norm wrt {\tt app} and {\tt input}. 

Let \( \Sigma \) be a \( \sigma \)-algebra over \( \M. \) The weight function in Figure \ref{fig:matrix factorization} determines a measure \( \mu \) such that for any \( E\in \Sigma \): \[
	\mu(E) = \inf \set{ {\texttt {err}}\su M \mid M\in E }.
\]
Therefore, \( \#(\phi, \mu) \) yields the lowest   \( \texttt {err} \) value;  the model \( M \) such that \( {\texttt {err} }\su M = \#(\phi, \mu) \) is one with the best factorization of matrix \( I. \)

\end{example}

\begin{figure}
	\begin{framed}\small\tt
		(set-logic LRA) \\\
		(set-algebra [REAL,inf,0]) \\
		(declare-function input (INT,INT) REAL) \\
		... /* declare left, right, app */ \\		
		(declare-function err () REAL) \\
		DATA = /* entries in input matrix (omitted) */\\
		F = app(x,y) == sum\{e\}  left(x,e)*right(e,y) \\
		G = err == norm(sum\{x,y\} input(x,y) - app(x,y)) \\
		(declare-weight TRUE err)\\
		(count (DATA and F and G))
	\end{framed} \caption{matrix factorization}\label{fig:matrix factorization}
\end{figure}

\subsection{Infinite: Factorized} %
\label{sub:counting_measure}

Despite the generality of the above definition, we would like to address the factorized setting for a number of applications, the most prominent being probabilistic inference in hybrid graphical models \cite{Belle:2015af}. Consider, for example, a joint distribution on the probability space \( \real\times \set{0,1} \). Here, it is natural to define weights for each random variable separately, prompting a factorized formulation of counting. More generally, in many robotic applications, such hybrid spaces are common \cite{thrun2005probabilistic}. The main idea then is to apply our definition for counting by measures to each variable independently, and construct a measure for the entire space by \emph{product measures} \cite{halmos-measure}.

A second technicality is that in the finite case, the set of literals true at a model was finite by definition. This is no longer the case. For example, suppose \( x \) is a real-valued variable in a language \( \L \), and \( M \) is a  \( \L \)-model that  assigns \( 3 \) to \( x. \) Then, \( M \models (x=3) \) but also \( M\models (x\neq 3.1), \) \( M\models (x\neq 3.11), \ldots, \) and so on. Thus, for technical reasons, we assume that \( \L \) only consists of constant symbols \( \X = \set{x\sub 1, \ldots, x\sub k} \) with fixed (possibly infinite) domains \( \set{D\sub 1, \ldots, D\sub k} \); the measures are defined for these domains.

\begin{definition}\label{defn measure amc} Let \( \AA = (\real, \oplus, \otimes, \idenz, \ideno) \) be any  semiring,   \(  \T = (\L, \M, \mm)\)  any theory where \( \sig(\L) = \set{x\sub 1, \ldots, x\sub k}  \) over fixed domains \( D\sub i. \) Suppose \( \phi\in \L. \) 
	Suppose \( \Sigma \sub i \) is a \( \sigma \)-algebra over \( D\sub i \) in that \( (D\sub i, \Sigma\sub i) \) is a \( \sigma \)-finite measurable space wrt the measure \( \mu\sub i\colon \Sigma\sub i \rightarrow \real \) respecting \( \oplus \) (as in Definition \ref{defn amc measure simple}).  Define the  product measure \( \mu\su * \doteq \mu\sub 1 \times \cdots \times \mu\sub k \) on the measurable space \( (D\sub 1 \times \cdots \times D\sub k, \Sigma\sub 1 \times \cdots \Sigma\sub k) \) satisfying:\footnote{The product measure is unique owing to the \( \sigma \)-finite assumption via the Hahn-Kolmogorov theorem \cite{halmos-measure}.}  \[
	\mu\su *(E\sub 1 \times \cdots \times E\sub k) = \mu\sub 1 (E\sub 1) \otimes \cdots \otimes \mu\sub k(E\sub k)
\]
for all \( E\sub i \in \Sigma\sub i. \) Finally,  define \[ \begin{array}{l}
	\amc(\phi, \mu\su *) = \mu\su *([\phi])
\end{array}
\]
where \(
	[\phi] =   \set{x\sub 1\su M \times \ldots \times x\sub k\su M \mid M \in \M(\phi) }.  \)

\end{definition}
Intuitively, \( E\sub 1\in \Sigma\sub 1, \ldots, E\sub k\in \Sigma\sub k \) capture sets of assignments, and the product measure considers the algebraic product of the weights on assignments to terms. As before, for \( E\sub i, E'\sub i \in \Sigma\sub i, \) \( \mu\sub i ( E\sub i \cup E'\sub i) = \mu\sub i (E\sub i) \oplus \mu\sub i (E'\sub i). \) Finally, precisely because the measures are defined on the domains of the terms, we obtain these  for all of the satisfying interpretations using the construction \( [\phi]. \)

\begin{example}\label{ex polyhedron} We demonstrate the problem of finding the volume of a polyhedron, needed in the static analysis of probabilistic programs \cite{sankaranarayanan2013static,RupakSMT}. Suppose \( \T = (\L, \M, \models) \) and \( \phi\in \L \) is as in Example \ref{ex:convex}, that is, \( \phi \) defines a polyhedron.  For every 0-ary function symbol \( x\sub i \in \set{x\sub 1, \ldots, x\sub k} \) with domain \( D\sub i = \real, \) let \( \Sigma\sub i \) be the set of all Borel subsets of \( \real, \) and let \( \mu\sub i \) be the Lebesgue measure. Thus, for any \( E\in \Sigma\sub i, \) \( \mu\sub i(E) \) gives the length of this line. Then, for the semiring \( (\real, +, \times, 0, 1) \), the \( + \) operator sums the lengths of lines for  each variable, and \( \times \) computes the products of these lengths. Thus, \( \amc(\phi,\mu\su *)  \) is the volume of \( \phi \).

	To see this in action, suppose \( \phi = (2x\leq 5) \land (x\geq 1) \land (0\leq y \leq 2). \) Then \( \M(\phi) = \set{\la x\rightarrow n, y \rightarrow m \ra \mid n, m \in \real, \models \phi\su{x,y}\sub {n,m}} \), that is, all assignments to \( x \) and \( y \) such that \( \phi\su{x,y}\sub {n,m} \) is a valid expression in arithmetic.  Therefore, 
\[	[\phi] = \set{(n,m) \mid n\in [1,2.5], m \in [0,2], n \in \real, m\in \real}. \]
Assuming \( \mu\sub x \) is the Lebesgue measure for all Borel subsets of the domain of \( x, \) we have \[ \mu\sub x(\set{n \mid n\in [1,2.5], n\in \real}) = 1.5. \]
Analogously, \( \mu\sub y (\set{m\mid m\in [0,2], m\in \real}) = 2. \) Then, the volume is  \( \#(\phi,\mu\su *) =  \mu\su *([\phi]) =   1.5 \times 2 = 3. \) 

\end{example}

\begin{example} We demonstrate probabilistic inference in hybrid models \cite{Belle:2015af} by extending Example \ref{ex polyhedron}. Consider a probabilistic program:  \[
	\begin{array}{l}
		X \sim \textsc{UNIFORM}(0,1) \\
		Y \sim \textsc{FLIP}(0,1)\\
		\textbf{if}~(X\gt .6 \textsc{ AND } Y\neq 1)~\textbf{then}~\textrm{return}~\textsc{DONE}
	\end{array}
\]
In English: \( X \) is drawn uniformly from [0,1] and \( Y \in \set{0,1} \) is the outcome of a coin toss. If \( X\gt .6 \) and \( Y \) is not 1, the program terminates successfully. Suppose we are interested in the probability of DONE,  which is expressed as the formula: \[
	\phi = (0\leq X\leq 1) \land (Y = 0 \lor Y = 1) \land (X\gt .6 \land Y \neq 1).
\] Suppose \( \T = (\L, \M, \models) \) is the theory of linear real arithmetic, with \( \sig(\L) = \set{X,Y} \), \( D\sub X =\real \) and \( D\sub Y = \set{0,1}. \) As in Example \ref{ex polyhedron}, let \( \Sigma\sub X \) be the set of all Borel subsets of \( \real \) and \( \mu\sub X \) the Lebesgue measure. Let \( \Sigma\sub Y \) be the set of all subsets of \( \set{0,1} \) and \( \mu\sub Y \) be the counting measure, \eg \( \mu\sub Y(\set{}) = 0, \mu\sub Y(\set{1}) = 1,  \) and $\mu\sub Y(\set{0,1}) = 2$. So \( \M(\phi) = \set{\la X\rightarrow n, Y \rightarrow m\ra \mid n\in \real, m\in \set{0,1}, \models \phi\su {X,Y}\sub {n,m}} \) and: \[\begin{array}{l}
	[\phi] = \set{(n,m) \mid n\gt .6,  0\leq n \leq 1, n\in \real, m\in \set{0,1}} \cap \\
	\quad\quad\quad \set{(n,m) \mid 0\leq n\leq 1, n\in \real, m\neq 1, m\in \set{0,1}}.
\end{array}
	\]
This means that \( \mu\sub X(\set{n\mid \exists m~~(n,m)\in [\phi]}) = .4 \) and also \( \mu\sub Y(\set{m\mid \exists n~~(n,m)\in \phi}) = 1 \). Thus, \( \mu\su *([\phi]) = .4\times 1. \) Analogously, \( \mu\su *([\phi\lor\neg\phi]) = 2. \) Therefore, the probability of DONE is \( .4/2 = .2. \)

\end{example}
In general, we have a variant of Proposition \ref{thm math programming} \cite{1384411}: 

\begin{theorem} Suppose \( \set{x\sub 1, \ldots, x\sub k} \) is any set of real-valued variables. Suppose \( {D} \) is any countable set. Suppose \( P \subset \real\su m \times {D}\su n \), where  \( m+n=k, \) is any region given by conjunctions of expressions of the form  \( c\sub 1 x\sub 1 + \ldots + c\sub k x\sub k \circ e \) and \( x\sub i \diamond d \), where \( \circ \in \set{\leq, \lt, \geq, \gt} \), \( \diamond\in\set{\neq, =} \),  \( c\sub i, e \in \real \) and \( d\in{D}. \) Then there is a \( \T = (\L, \M, \models), \AA, \mu \) and \( \phi\in \L \) such that \( \#(\phi,\mu) \) is the volume of \( P. \)

\end{theorem}

\section{Towards Compositionality} %
\label{sec:combining_theories}

A noteworthy feature of many logic-based knowledge representation formalisms is their compositional nature. In semiring programming, using the expressiveness of predicate logic, it is fairly straightforward  to combine theories over possibly different signatures (\eg propositional logic and linear arithmetic), as seen, for example, in SMT solvers \cite{BSST09HBSAT}. 

A more intricate flavor of compositionality is when the new specification becomes difficult (or impossible) to define using the original components. This is a common occurrence in large software repositories, and has received a lot of attention in the AI community  \cite{DBLP:conf/aaai/LierlerT15}. 

In this section, we do not attempt to duplicate such efforts, but propose a different account of compositionality that is closer in spirit to semiring programming. It builds on similar ideas for CSPs \cite{Bistarelli:2004:SSC:993462}, and is motivated by machine learning problems where learning (\ie optimization) and inference (\ie model counting) need to be addressed in tandem. More generally, the  contribution here allows us to combine two semiring programs, possibly
 involving  different semirings. For simplicity of presentation, we consider non-factorized and finite problems over distinct vocabularies.  

\begin{definition} Suppose \( \AA\sub 1 = (\A\sub 1, \oplus\sub 1, \idenz\sub 1) \) and \( \AA\sub 2 = (\A\sub 2, \oplus\sub 2, \idenz\sub 2) \) are any two semirings. We define their \emph{composition} \( \AA = (\A, \oplus, \idenz) \) as: \begin{itemize}
	\item \( \A = \set{(a,b) \mid a\in \A\sub 1, b\in \A\sub 2} \);
	\item \( \idenz = (\idenz\sub 1, \idenz\sub 2) \);
	\item for every \( c = (a,b) \in \A \) and \( c' = (a',b') \in \A \), let \( c \oplus c' = (a\oplus\sub 1 a', b\oplus\sub 2 b') \).
\end{itemize}
	
\end{definition}

That is, the composition of the semirings is formed from the Cartesian product, respecting the summation operator for the individual structures. 

\begin{definition}\label{defn theory compose} Suppose \( \T\sub 1 = (\L\sub 1, \M\sub 1, \mm\sub 1) \) and \( \T\sub 2 = (\L\sub 2, \M\sub 2, \mm\sub 2) \) are theories, where \( \L\sub 1 \) and \( \L\sub 2 \) do not share atoms, 
	\( \AA\sub 1 \) and \( \AA\sub 2 \) semirings, and \( w\sub 1 \) and \( w\sub 2 \) weight functions for \( \M\sub 1 \) and \( \M\sub 2 \) respectively. Given the environments \( (\T\sub 1, \AA\sub 1, w\sub 1) \) and \( (\T\sub 2, \AA\sub 2, w\sub 2) \), we define its \emph{composition} as \( (\T, \AA, w) \), where  \( \AA \) is a composition of \( \AA\sub 1 \) and \( \AA\sub 2 \) and \( \T = (\L, \M, \models) \):  \begin{itemize}
	\item   \( \phi\in \L \) is obtained over Boolean connectives from \( \L\sub 1 \cup \L\sub 2 \).  
	\item \( \M = \set{(M\sub 1, M\sub 2) \mid M\sub i \in \M\sub i }. \)
	\item The meaning of \( \phi\in \L \) is defined inductively: \begin{itemize}
	\item \( (M\sub 1, M\sub 2) \models p   \) for atom \( p \) iff \( M\sub 1 \mm\sub 1 p \) if \( p\in \L\sub 1  \) and \( M\sub 2 \mm\sub 2 p \) otherwise; 
	\item \( (M\sub 1, M\sub 2) \models \neg \phi \) iff not the case that \( (M\sub 1, M\sub 2) \models \phi \);
	\item \( (M\sub 1, M\sub 2) \models  \phi\sub x \land \phi\sub y \) iff \( (M\sub 1,M\sub 2) \models \phi\sub i \) for \( i\in\set{x,y} \). 
\end{itemize}
\item For any \( (M\sub 1, M\sub 2) \in\M, \)   \( w((M\sub 1,M\sub 2)) = (w\sub 1(M\sub 1), w\sub 2(M\sub 2))  \). 
\end{itemize}
	
\end{definition}

In essence, the Cartesian product for the semirings is extended for arbitrary theories and weight functions. The meaning of formulas rests on the property that \( \L\sub 1 \) and \( \L\sub 2 \) do not share atoms.\footnote{This is analogous to SMT solvers for combinations of theories  \cite{BSST09HBSAT}.  However, see  \cite{DBLP:conf/aaai/LierlerT15,DBLP:conf/ijcai/EnsanT15} for accounts of modularity based on first-order structures sharing vocabularies (and thus, atoms).} It is now easy to see that the counting for problem for \( \T \) works as usual: that is, for any \( \phi\in \L \), \[
	\#(\phi,w) = \bigoplus\sub {(M\sub 1, M\sub 2) \models \phi} w((M\sub 1, M\sub 2)).
\]

\begin{example} We demonstrate a (simple) instance of combined learning and inference. Imagine a robot navigating a world by performing \emph{move} actions, and believes its actuators need repairs. But before it alerts the technician, it would  like to test this belief. A reasonable test, then, is to inspect its trajectory so far, and check whether the expected outcome of a move action in the current state matches the behavior of the very first move action. More precisely, the robot needs to appeal to linear regression to estimate its expected outcome, and query its beliefs based on the regression model. We proceed as follows.

\subsection*{Environment 1} %
\label{ssub:environment_1} Let \( \T\sub 1 = (\L\sub 1, \M\sub 1, \models\sub 1) \) be the theory of real arithmetic with \( \sig(\L\sub 1) = \set{s\sub 0, s\sub 1, \ldots, s\sub k, a, b,e} \). Suppose that by performing a \emph{move} action, the robot's position changes from \( s\sub i \) to \( s\sub {i+1}. \) Let \( \phi\sub 1 \in\L\sub 1 \) be as follows: \[\begin{array}{l}
 (s\sub 0 = 1 \land s\sub 1 = 2 \land s\sub 2 = 3) ~\land \\
	e = \sum\sub {i} (s\sub {i+1} - b -a \cdot s\sub i)\su 2
\end{array}
\]
The idea is that the values of \( s\sub i \) are the explanatory variables in the regression model and \( s\sub {i+1} \) are the response variables, that is, the trajectory data is  of the form \( \set{(s\sub 0, s\sub 1), (s\sub 1, s\sub 2)} \). 

Consider a  weight function \( w\sub 1\colon M \rightarrow e\su M. \) That is, the weight of a model \( M \) is the universe element corresponding to the constant \( e \) in \( M \), analogous to Figure \ref{fig:logic mathemtical programming}. (For simplicity, we assume that the coefficients of the regression model are natural numbers.) 
For the semiring \( \AA\sub 1 = (\nat, \min, \infty) \): \[
	\#(\phi\sub 1,w\sub 1) = \min \set{e\su M \mid M\models \phi\sub 1, M\in \M\sub 1}.
\]
In other words, models in \( \M\sub 1 \) interpret \( s\sub i \) as given by the data,  and models differ in their interpretation of \( a, b \) and thus, \( e. \) For the data in \( \phi\sub 1, \) we would have a model \( M \) where \( e\su M = 0 \), \( a\su M = 1 \) and \( b\su M = 1 \), and so  \( \#(\phi\sub 1,w\sub 1) =0. \)

\subsection*{Environment 2} %
\label{ssub:environment_2}
Let \( \T \sub 2 = (\L\sub 2, \M\sub 2, \models\sub 2) \) be a propositional theory, where \( \sig(\L\sub 2) = \set{{\it repair}}. \) Imagine a weight function \( w\sub 2 \) as follows: \( w\sub 2(\set{{\it repair}}) = .7,\) and \(  w\sub 2(\set{\neg{\it repair}}) = .3.\) That is, the robot believes that repairs are needed with a higher probability. Indeed, for the semiring \( (\real,+,0) \): \[
	\#({\it repair},w\sub 2) = .7 \textrm{ versus } \#(\neg {\it repair},w\sub 2) = .3.
\]

\subsection*{Composition} %
\label{ssub:composition}
Let \( (\T, \AA, w) \) be the composition of the two environments with \( \T = (\L, \M, \models) \) and \( \AA = (\A, \oplus, \idenz) \). Suppose \( \phi \in \L \) is as follows: \[
	\begin{array}{l}
		\phi\sub 1 \land  
		(s\sub 3 = a\cdot s\sub 2 +b) ~\land \\
		(s\sub 3 - s\sub 2) \neq (s\sub 1 - s\sub 0) \equiv {\it repair}. 
	\end{array}
\]
The final conjunct basically checks whether the expected change in position matches what was happening initially, and if not, \emph{repair} should be true. 

To see Definition \ref{defn theory compose} in action, observe that for any \( M\sub 1\in \M\sub 1, M\sub 2 \in \M\sub 2, \)  \( (M\sub 1, M\sub 2) \in \M \), the formulas \( \phi\sub 1 \) and those involving \( s\sub i \) 
 are interpreted in \( M\sub 1, \) but {\it repair} in \( M\sub 2. \)  The weight function is as follows. Given \( (M\sub 1, M\sub 2) \in \M \) and \( (M'\sub 1, M'\sub 2)\in \M \), we have: \[ \begin{array}{l}
 	w((M\sub 1, M\sub 2)) \oplus w((M'\sub 1, M'\sub 2)) \\ 
	\quad \quad  = (~\min(w\sub 1(M\sub 1), w\sub 1(M'\sub 1)),~w\sub 2(M\sub 2) + w\sub 2(M'\sub 2) ~).
 \end{array} \]
Then the robot can obtain the weight of
 {\it repair} and  \( \phi \) using: \[ \#(\phi\land {\it repair},w) = (0,0) \]
where, of course, the first argument is the error of the regression model and the second is 0 because \( \phi\land {\it repair} \) is inconsistent. That  can be contrasted to the count below: \[
	\#(\phi \land \neg{\it repair},w) = (0,.3).
\] It is also easy to see  that \(
	\#(\phi,w) = (0,.3).
\)

As in WMC \cite{DBLP:journals/ai/ChaviraD08}, suppose the robot obtains the probability of
a query \( q \) given \( \phi \) using: \[
	\frac{\#(\phi\land q, w)}{\#(\phi,w)}
\]
where the division is carried out by ignoring the regression error. Then the probability of \( \neg {\it repair} \) given \( \phi \) is 1. Thus, no repairs are needed. 

\end{example}

\begin{example} We show  that by combining theories, a natural semantics can be given to hybrid problems such as task and motion planning. 
	%
	Here, the concern is to integrate a symbolic high-level planner, described  in logic, and geometric constraints; see   \cite{DBLP:conf/icra/SrivastavaFRCRA14} for terminology and notation.  Suppose \( \Z \) is the theory of  integer  arithmetic that interprets a motion planning space \( (C, N, p) \) where \( C = \mathbb{Z} \) is the configuration space, \( N \subset C \) is an obstacle region, and \( p\in C \) is the initial pose of a robot's gripper. (For simplicity, we consider a one-dimensional setting.) The result of doing a motion \( t \) is the pose \( p+t \). We  axiomatize that \( x \) is reachable from \( y \) by following \( t \) without touching obstacles as: \[ \textit{IsReachable(x,y,t)} \equiv (x = y + t) \land \forall z (z\leq t \supset (y+z \not\in N)). \]	 
	
	Next, suppose \( \T \) is a propositional theory that interprets a task planning space \( (S, A, i, g) \), where \( A \) is a set of actions, \( S \) a set of states, and \( i, g \in S \) are the initial and goal states. The task is to synthesize action sequences reaching \( g. \) 
	 However, in robotic applications, actions are predicated on  geometric constraints; \eg the precondition for  \emph{pickup(o,p,t)} -- the action of picking up an object \( o \) wrt a  trajectory \( t \) and the gripper's pose \( p \) --  is the sentence: \[ \emph{IsReachable(o,p,t)} \land  \emph{IsGripperFree}.\] Interestingly,  \emph{IsGripperFree} is from the language of \( \T \) but \emph{IsReachable(x,y,z)} is from that of \( \Z \), and using our semantic setup, such complex   systems can  be  easily interpreted. 
	 
	 To describe an illustrative counting problem, suppose that for every \( \Z \)-model \( M \) and \( \T \)-model \( M', \) \( (M,M') \) determines a combined task-motion plan of some length.  A plan is valid iff \( g \) can be reached from \( i \) by following this plan wrt the domain's axioms~(\eg avoiding obstacle regions  and satisfying action preconditions). Suppose plans are of the form \( t\sub 1, a\sub 1, \ldots, t\sub m, a\sub n  \), where \( a\sub i \in A \) and \( t\sub j \) are motions. 
	 	 Suppose actions and motions incur costs, and the weight of a model is \( \sum\sub i \emph{cost}(a\sub i) + \sum\sub j \emph{cost}(t\sub j) \) wrt the plan it determines. Assuming \( \Z\cup \T \) is a finite theory (for simplicity), the semiring  \( (\real, \min, +, 0, 1) \) yields a cost optimal plan.\footnote{Prior work on semiring composition \cite{Bistarelli:2004:SSC:993462} can further allow us to minimize one aspect (\eg cost) and maximize  another (\eg number of rooms cleaned).}

\end{example}

\section{Solver Construction} %
\label{sec:solver_strategy} 

The upshot of semiring programming is that it encourages us to inspect  strategies for a unified inferential mechanism \cite{kautzunified}. This has to be done carefully, as we would like to build on scalable methodologies in the literature, by restricting logical theories where necessary. In this section, we discuss whether our programming model can be made to work well in practice. 

Let us  consider two extremes: \begin{itemize}
	\item \textbf{Option 1:}  At one extreme is a solver strategy based on a  single computational technique. Probabilistic programming languages, such as Church \cite{conf/uai/GoodmanMRBT08}, have made significant progress in that respect for generative stochastic processes by appealing to Markov Chain Monte Carlo sampling techniques. Unfortunately, such sampling techniques do not scale well on large problems and have little support for linear and logical constraints. 
	\item \textbf{Option 2:} At the other extreme is a solver strategy that is arbitrarily heterogeneous, where we develop unique solvers for specific  environments, that is, \( (\T, \S) \) pairs.  
\end{itemize}


\textbf{Option 3:} 
We believe the most interesting option is in between these two extremes. In other words, to identify the smallest set of computational techniques, and effectively integrate them is both challenging and insightful. This may mean that such a strategy is less optimal than Option 2 for the environment, but we would obtain a simpler and more compact execution model. To that end, let us make the following observations from our inventory of examples: \begin{itemize}
	\item \textbf{Finite versus infinite:} variable assignments are taken from finite sets versus infinite or uncountable sets. 
	\item \textbf{Non-factorized versus factorized:} the former is usually an optimization problem with an objective function that is to be maximized or minimized. The latter is usually a counting problem, where we would need to identify one or all solutions. 
	\item \textbf{Compositionality:} locally consistent solutions (\ie in each environment \( (\T,\AA) \)) need to be tested iteratively for global consistency. 
\end{itemize}
Thus, Option 3 would be realized as follows: \begin{itemize}
	\item Factorized problems need a methodology for effective enumeration, and  therefore, advances in model counting~\cite{modelcountingchapter}, such as knowledge compilation, are the most relevant. For finite theories, we take our cue from the Problog family of languages \cite{DBLP:conf/uai/FierensBTGR11,DBLP:conf/aaai/KimmigBR11}, that have effectively applied arithmetic circuits for tasks such as WMC and MPE. In particular, it is shown in \cite{AMC} how arbitrary semiring labels can be propagated in the circuit. See \cite{fargier2013towards,DBLP:conf/ijcai/FargierMS13} for progress on knowledge compilation in CSP-like environments.   

For infinite theories, there is growing interest in effective model counting for linear arithmetic using SMT technology  \cite{RupakSMT,Belle:2015af,Belle:2016aaai}. Like in \cite{DBLP:conf/aaai/KimmigBR11}, however, we would need to extend these counting approaches to arbitrary semirings. 

\item For non-factorized problems, a natural candidate for handling semirings does not immediately present itself, making this is a worthwhile research direction.\footnote{Approaches like \cite{DBLP:journals/corr/BerreLM14,DBLP:conf/ijcai/FargierMS13} on knowledge compilation for optimization and \cite{DBLP:conf/ijcai/SannerM05} on circuits for linear constraints are  compelling, but their applicability to non-trivial arithmetic and optimization problems remains to be explored.} Appealing to off-the-shelf optimization software \cite{andrei2013nonlinear} is always an option, but they embody diverse techniques and the absence of a simple high-level solver strategy makes adapting them for our purposes less  obvious. In that regard, solvers for \emph{optimization modulo theories} (OMT) \cite{Sebastiani:2015:OMT:2737801.2699915} are perhaps the most promising. OMT technology extends SMT technology in additionally including a cost function that is be maximized (minimized). 
In terms of expressiveness, CSPs \cite{nieuwenhuis2006sat} and 
certain classes of mathematical programs can be expressed, even in the presence of logical connectives. In terms of a solver strategy, they use binary search in tandem with  lower and upper bounds to find the maximum (minimum). This is not unlike DPLL traces in knowledge compilation, which makes that technology the most accessible for propagating arbitrary semiring labels.

\item Compositional settings are, of course, more intricate. Along with OMT, and classical iterative methods  like \emph{expectation maximization} \cite{books/daglib/0023091}, there are a number of recent approaches employing branch-and-bound search strategies to navigate between local and global consistency \cite{Friesen:2015aa}. Which of these can be made amenable to compositions of SP programs remains to be seen however. \smallskip 

\end{itemize}
Overall, we believe the most promising first step is to limit the vocabulary of the logical language to propositions and constants (\ie 0-ary functions), which make appealing to knowledge compilation and OMT technology straightforward. It will also help us better characterize the complexity of the problems that SP attempts to solve.\footnote{Richer fragments have  to be considered carefully to avoid undecidable properties  \cite{boerger1997classical}.} At first glance, SP is seen to naturally capture \#P-complete problems in the factorized setting, both in the finite case \cite{modelcountingchapter} and the infinite one~\cite{journals/siamcomp/DyerF88}. 
In the non-factorized setting, many results from OMT and mathematical programming are inherited depending on the nature of the objective function and the domains of the program variables \cite{kannan1978computational,BSST09HBSAT,Sebastiani:2015:OMT:2737801.2699915}. By restricting the language as suggested, the applicability of these results can be explored more thoroughly.

\section{Related Work} %
\label{sec:related_approaches} 

Semiring programming  is related to efforts from different disciplines within AI, and we discuss  representative camps. In a nutshell, SP  can be seen as a very general semantical framework, as noted  in Figure \ref{fig:rel}. 

\begin{figure}[t]
	\begin{center} \small
		\begin{tabular}{l|c|c|c|c|c|c|c}
		              & FT & FN & IT & IN & C & S & R \\
		\hline
		{\sc AProblog} & $\checkmark$ &  \( \times \) & \( \times \) & \( \times \) &  \( \checkmark \) & \( \checkmark \) & $\checkmark\su \bullet$ \\
		{\sc Church} & \( \checkmark \) & \( \times \) & \( \checkmark \) & \( \times \) & \( \checkmark\su \bullet \) & \( \times \) & \( \times \) \\
		{\sc Essence} & \( \times \) & \( \checkmark \) & \( \times \)  & \( \times \) & \( \checkmark\su \bullet \) & \( \times \) & \( \times \) \\
{\sc CSP} & \( \times \) & \( \checkmark\su\bullet \) & \( \times \) & \( \checkmark\su\bullet \) & \( \checkmark\su \bullet \) & \( \times \) & \( \times \) \\ 
{\sc Semiring CSP} & \( \times \) & \( \checkmark\su\bullet \) & \( \times \) & \( \times \) & \( \checkmark\su \bullet \) & \( \checkmark \) & \( \times \) \\
{\sc AMPL} & \( \times \) & \( \checkmark \) & \( \times \) & \( \checkmark \) & \( \checkmark\su \bullet \) & \( \times \) & \( \times \) \\
{\sc ASP} & \( \checkmark \) & \( \checkmark\su \bullet \) & \( \times \) & \( \times \) & \( \checkmark \) & \( \times \) & \( \checkmark \) \\
{\sc (O)SMT} & \( \times \) & \( \checkmark \) & \( \times \) & \( \checkmark \) & \( \checkmark \) & \( \times \) & \( \times \) \\
{\sc \#SMT} & \( \checkmark \) & \( \times \) & \( \checkmark \) & \( \times \) & \( \checkmark \) & \( \times \)  & \( \times \) \\
{\sc SP} & \( \checkmark \)  & \( \checkmark \) & \( \checkmark \) & \( \checkmark \) & \( \checkmark \) & \( \checkmark \) & \( \checkmark \)
		\end{tabular}
	\end{center}
		\caption{\footnotesize A comparison, where F = finite, T = factorized, N = non-factorized, I = infinite, C = logical connectives and quantifiers, R = non-standard, S = semiring apparatus, and \( \bullet \) denotes that a feature is available in a restricted sense.} \label{fig:rel}
\end{figure}

\subsection{Statistical modeling} %
\label{ssub:machine_learning} 
Formal languages for generative stochastic processes, such as Church \cite{conf/uai/GoodmanMRBT08} and BLOG \cite{DBLP:conf/ijcai/MilchMRSOK05},  have received a lot of attention in the learning community. Such languages  provide mechanisms to compactly specify complex probability distributions, and appeal to sampling for inference.

Closely related to such proposals are probabilistic logic programming languages such as Problog \cite{DBLP:conf/ijcai/RaedtKT07} that extends Prolog with probabilistic choices and uses WMC for inference \cite{DBLP:conf/uai/FierensBTGR11}. In particular, a semiring generalization of Problog, called aProbLog \cite{DBLP:conf/aaai/KimmigBR11}, was the starting point for our work and employs the semiring variation 
of WMC for inference \cite{AMC}. A recent extension of aProbLog, called kProbLog,  by \cite{Orsini2017} 
is able to further combine several semirings and  does not require factorized weights as it uses meta-functions $w(a) = f(w(a_1), ... ,w(a_n))$ to compute the weight $w(a)$ of an atom from the weights of the atoms $a_1, ... , a_n$ appearing in its proofs. But the kind of weight function and factorization used in kProbLog differs from the unfactorized weight function over the models used in the present paper. Nevertheless, kProbLog is able to represent tensors, compute kernel functions and perform algorithmic differentiation. It will be interesting to investigate whether kProbLog can be combined with semiring programming. 



As discussed before, SP generalized the formulation of algebraic model counting (AMC) \cite{AMC}, and in that sense, provides a semantic characterization for the sum product problem \cite{DBLP:journals/jair/BacchusDP09} to richer class of languages and models. But by giving up  the propositional language, and in particular, the use of arithmetic circuits, we loose the tractability results and unified evaluation scheme offered by AMC. Of course, it is always possible to restrict and/or otherwise map infinitary languages to finite and decomposable grammars via abstraction. This has been investigated in the case of a recent continuous extension to WMC called weighted model integration \cite{Belle:2015af,Belle:2016aaai}, where by interpreting the pieces of a density function as propositions, propositional circuits are leveraged for inference. In an effort independent to ours\footnote{A preliminary version of this work was online on September 2016: https://arxiv.org/abs/1609.06954}, it is shown how an algebraic extension to sum product networks \cite{poon2011sum} enable tractable problem solving  \cite{friesen2016sum}, closely following the observations in \cite{AMC}. In that work, a  continuous extension is considered as well, but under the assumption that the specification of the weight function as well as the computation of the count can be factorized.

The use of semirings in machine learning is not new to aProbLog, see \eg  \cite{DBLP:journals/coling/Goodman99}, and programming languages such as Dyna \cite{eisner-filardo-2011}. Dyna is based on Datalog; our logical setting is strictly more expressive than Datalog and its extensions (\eg non-Horn fragment, constraints over reals). Dyna also labels proofs but not interpretations, as would SP (thus   capturing weighted model counting, for example).

\subsection{Constraints} %
\label{ssub:constraints}
The constraints literature boasts a variety of modeling languages, such as Essence \cite{DBLP:journals/constraints/FrischHJHM08}, among others \cite{Marriott:2008aa,DBLP:journals/informs/Hentenryck02}. (See \cite{DBLP:conf/cp/FontaineMH13}, for example, for a proposal on combining heterogeneous  solvers.) On the one hand, SP is more expressive from a logical viewpoint as constraints can be described using arbitrary formulas from predicate logic, and we address many problems beyond constraints, such as probabilistic reasoning. On the other hand, such constraint  languages make it easier for non-experts to specify problems while SP, in its current form, assumes a background in logic. Such languages, then, would be of interest for extending SP's modeling features. 

A notable line of CSP research is by Bistarelli \cite{Bistarelli:2004:SSC:993462} and his colleagues \cite{DBLP:journals/constraints/BistarelliMRSVF99,Bistarelli:2004:SSC:993462}. Here, semirings are used for diverse CSP specifications, which has also been realized in a CLP framework  \cite{bistarelli2001semiring}. In particular, our account of compositionality is influenced by \cite{Bistarelli:2004:SSC:993462}.  Under some representational assumptions, SP and such accounts are related, but as noted, SP can formulate problems such as probabilistic inference in hybrid domains that does not have an obvious analogue in these accounts.

\subsection{Optimization} %
\label{ssub:optimization}
Closely related to the constraints literature are the techniques embodied in mathematical programming more generally. There are three major traditions in this literature that are related to SP. Modeling languages such as AMPL \cite{opac-b1123349} are fairly close to constraint modeling languages, and even allow parametrized constraints, which are ground at the time of search. The field of disciplined programming \cite{gb08} supports features such as object-oriented constraints. Finally, relational mathematical programming \cite{DBLP:conf/aaai/ApselKM14} attempts to exploit symmetries in parametrized constraints. 

From a solver construction perspective, these languages present interesting possibilities. From a framework point of view, however, there is little support for logical reasoning in a general way.

\subsection{Knowledge representation} %
\label{ssub:knowledge_representation}
Declarative problem solving is a focus  of many proposals, including ASP \cite{brewka2011answer}, model expansion \cite{DBLP:conf/aaai/MitchellT05,DBLP:conf/ijcai/TernovskaM09}, among others \cite{TLP:703704}. These proposals are (mostly) for problems in NP, and so do not capture \#P-hard problems like model counting and WMC. Indeed, the most glaring difference is the absence of weight functions over possible worlds, which is central to the formulation of  statistical models. Weighted extensions of these formalisms, \eg   \cite{DBLP:journals/tplp/BaralGR09,DBLP:conf/kr/LiuJN12}, are thus closer in spirit. 


The generality of SP also allows us to instantiate many such proposals,  including formalisms using linear arithmetic fragments   \cite{Belle:2015af,Sebastiani:2015:OMT:2737801.2699915}. Consider OMT for example. OMT can be used to express quantifier-free linear arithmetic sentences with a linear cost function, and a first-order structure that minimizes the cost function is sought. From a specification point of view, SP does not limit the logical language, does not require that objective functions be linear, and a variety of model comparisons, including counting, are possible via semirings. Compositionality in SP, moreover, goes quite beyond this technology.

Finally, there is  a  longstanding interest in combining different (logical) environments in a single logical framework, as seen, for example, in modular and multi-context systems  \cite{DBLP:conf/aaai/LierlerT15,DBLP:conf/ijcai/EnsanT15}. In such frameworks, it would be possible to get a ILP program and ASP program to communicate their solutions, often by sharing atoms. In our view, Section \ref{sec:combining_theories} and these frameworks emphasize different aspects of compositionality. The SP scheme assumes the modeler will formalize a convex optimization problem and a SAT problem in the same programming language since they presumably arise in a single application (\eg a task and motion planner); this allows model reuse and enables transparency. In contrast, modular systems essentially treat diverse environments as black-boxes, which is perhaps easier to realize. On the one hand, it would be interesting to see whether modular systems can address problems such as combined inference and learning. On the other hand, some applications may require that different  environments share atoms, for which our  account on compositionality could be extended by borrowing ideas from modular systems. 

\section{Conclusions}
\label{sec:discussion}
In a nutshell, SP is a framework to declaratively specify four major concerns in AI applications: \begin{itemize}
	\item logical reasoning; 
	\item non-standard models; 
	\item discrete and continuous probabilistic inference; 
	\item discrete and continuous optimization. 
\end{itemize}
To its strengths, we find that SP is {\em universal} (in the above sense) and \emph{generic} (in terms of allowing instantiations to particular   logical languages and semirings). Thus, we believe SP represents a simple, uniform, modular and transparent approach to the model building process of complex AI applications. 

SP comes with a rigorous semantics to give meaning to its programs. In that sense, we imagine future developments of SP would follow constraint programming languages and probabilistic programming languages in providing more intricate modeling features which, in the end, resort to the proposed semantics in the paper. 

Perhaps the most significant aspect of SP is that it also allows us to go beyond existing paradigms as the richness  of the framework admits novel formulations that combine 
theories from these different fields, as illustrated by means of a combined  regression and probabilistic inference example. In the long term, we hope SP will contribute to the bridge between learning and reasoning.

%
%
%
%
%
%
%
%
%

%

%
%
%
%
%
%
%
%
%
%
%
%
%
%
%
%
%
%

 %
 %
 %
 %

%
%

%

%

%
%
%

%

%
%
%
%
%

 %
 %
 %

 %

%
%
%

%

\bibliographystyle{abbrv}

\begin{thebibliography}{10}

\bibitem{andrei2013nonlinear}
N.~Andrei.
\newblock {\em Nonlinear Optimization Applications Using the GAMS Technology}.
\newblock Springer, 2013.

\bibitem{DBLP:conf/aaai/ApselKM14}
U.~Apsel, K.~Kersting, and M.~Mladenov.
\newblock Lifting relational map-lps using cluster signatures.
\newblock In {\em AAAI}, pages 2403--2409, 2014.

\bibitem{aziz2015stable}
R.~A. Aziz, G.~Chu, C.~Muise, and P.~J. Stuckey.
\newblock Stable model counting and its application in probabilistic logic
  programming.
\newblock In {\em Twenty-Ninth AAAI Conference on Artificial Intelligence},
  2015.

\bibitem{DBLP:journals/jair/BacchusDP09}
F.~Bacchus, S.~Dalmao, and T.~Pitassi.
\newblock Solving {\#}{SAT} and {B}ayesian inference with backtracking search.
\newblock {\em J. Artif. Intell. Res.}, 34:391--442, 2009.

\bibitem{DBLP:journals/tplp/BaralGR09}
C.~Baral, M.~Gelfond, and J.~N. Rushton.
\newblock Probabilistic reasoning with answer sets.
\newblock {\em TPLP}, 9(1):57--144, 2009.

\bibitem{baras2010path}
J.~S. Baras and G.~Theodorakopoulos.
\newblock Path problems in networks.
\newblock {\em Synthesis Lectures on Communication Networks}, 3(1):1--77, 2010.

\bibitem{BarFT-RR-15}
C.~Barrett, P.~Fontaine, and C.~Tinelli.
\newblock {The SMT-LIB Standard: Version 2.5}.
\newblock Technical report, Department of Computer Science, The University of
  Iowa, 2015.
\newblock Available at {\tt www.SMT-LIB.org}.

\bibitem{BSST09HBSAT}
C.~Barrett, R.~Sebastiani, S.~A. Seshia, and C.~Tinelli.
\newblock Satisfiability modulo theories.
\newblock In {\em Handbook of Satisfiability}, chapter~26, pages 825--885. IOS
  Press, 2009.

\bibitem{BELLE2018189}
V.~Belle and H.~J. Levesque.
\newblock Reasoning about discrete and continuous noisy sensors and effectors
  in dynamical systems.
\newblock {\em Artificial Intelligence}, 262:189 -- 221, 2018.

\bibitem{Belle:2015af}
V.~Belle, A.~Passerini, and G.~Van~den Broeck.
\newblock Probabilistic inference in hybrid domains by weighted model
  integration.
\newblock In {\em IJCAI}, 2015.

\bibitem{Belle:2016aaai}
V.~Belle, A.~Passerini, and G.~Van~den Broeck.
\newblock Towards component caching in hybrid domains.
\newblock In {\em AAAI}, 2016.

\bibitem{DBLP:journals/corr/BerreLM14}
D.~L. Berre, E.~Lonca, and P.~Marquis.
\newblock On the complexity of optimization problems based on compiled {NNF}
  representations.
\newblock {\em CoRR}, abs/1410.6690, 2014.

\bibitem{Bistarelli:2004:SSC:993462}
S.~Bistarelli.
\newblock {\em Semirings for Soft Constraint Solving and Programming}.
\newblock SpringerVerlag, 2004.

\bibitem{bistarelli2001semiring}
S.~Bistarelli, U.~Montanari, and F.~Rossi.
\newblock Semiring-based constraint logic programming: syntax and semantics.
\newblock {\em {TOPLAS}}, 23(1):1--29, 2001.

\bibitem{DBLP:journals/constraints/BistarelliMRSVF99}
S.~Bistarelli, U.~Montanari, F.~Rossi, T.~Schiex, G.~Verfaillie, and
  H.~Fargier.
\newblock Semiring-based {CSP}s and valued {CSP}s: Frameworks, properties, and
  comparison.
\newblock {\em Constraints}, 4(3):199--240, 1999.

\bibitem{boerger1997classical}
E.~Boerger, E.~Gr{\"a}del, and Y.~Gurevich.
\newblock {\em The classical decision problem}.
\newblock Springer Verlag, 1997.

\bibitem{brewka2011answer}
G.~Brewka, T.~Eiter, and M.~Truszczy{\'n}ski.
\newblock Answer set programming at a glance.
\newblock {\em Communications of the ACM}, 54(12):92--103, 2011.

\bibitem{TLP:703704}
M.~Cadoli and T.~Mancini.
\newblock Combining relational algebra, sql, constraint modelling, and local
  search.
\newblock {\em Theory and Practice of Logic Programming}, 7:37--65, 1 2007.

\bibitem{DBLP:journals/ai/ChaviraD08}
M.~Chavira and A.~Darwiche.
\newblock On probabilistic inference by weighted model counting.
\newblock {\em Artif. Intell.}, 172(6-7):772--799, 2008.

\bibitem{RupakSMT}
D.~Chistikov, R.~Dimitrova, and R.~Majumdar.
\newblock Approximate counting in smt and value estimation for probabilistic
  programs.
\newblock In {\em TACAS}, volume 9035, pages 320--334. 2015.

\bibitem{DBLP:conf/ijcai/RaedtKT07}
L.~De~Raedt, A.~Kimmig, and H.~Toivonen.
\newblock Problog: {A} probabilistic prolog and its application in link
  discovery.
\newblock In {\em Proc. IJCAI}, pages 2462--2467, 2007.

\bibitem{Ding:2010:CSM:1687044.1687110}
C.~H.~Q. Ding, T.~Li, and M.~I. Jordan.
\newblock Convex and semi-nonnegative matrix factorizations.
\newblock {\em IEEE Trans. Pattern Anal. Mach. Intell.}, 32(1):45--55, Jan.
  2010.

\bibitem{journals/siamcomp/DyerF88}
M.~E. Dyer and A.~M. Frieze.
\newblock On the complexity of computing the volume of a polyhedron.
\newblock {\em SIAM J. Comput.}, 17(5):967--974, 1988.

\bibitem{eisner-filardo-2011}
J.~Eisner and N.~W. Filardo.
\newblock Dyna: Extending {D}atalog for modern {AI}.
\newblock In {\em Datalog Reloaded}, LNCS, pages 181--220. Springer, 2011.

\bibitem{enderton1972mathematical}
H.~Enderton.
\newblock {\em A mathematical introduction to logic}.
\newblock Academic press New York, 1972.

\bibitem{DBLP:conf/ijcai/EnsanT15}
A.~Ensan and E.~Ternovska.
\newblock Modular systems with preferences.
\newblock In {\em IJCAI}, pages 2940--2947, 2015.

\bibitem{fargier2013towards}
H.~Fargier, P.~Marquis, and A.~Niveau.
\newblock Towards a knowledge compilation map for heterogeneous representation
  languages.
\newblock In {\em IJCAI}, pages 877--883, 2013.

\bibitem{DBLP:conf/ijcai/FargierMS13}
H.~Fargier, P.~Marquis, and N.~Schmidt.
\newblock Semiring labelled decision diagrams, revisited: Canonicity and
  spatial efficiency issues.
\newblock In {\em IJCAI}, 2013.

\bibitem{DBLP:conf/uai/FierensBTGR11}
D.~Fierens, G.~Van~den Broeck, I.~Thon, B.~Gutmann, and L.~De~Raedt.
\newblock Inference in probabilistic logic programs using weighted {CNF}'s.
\newblock In {\em UAI}, pages 211--220, 2011.

\bibitem{DBLP:conf/cp/FontaineMH13}
D.~Fontaine, L.~Michel, and P.~Van~Hentenryck.
\newblock Model combinators for hybrid optimization.
\newblock In {\em CP}, pages 299--314, 2013.

\bibitem{opac-b1123349}
R.~Fourer, D.~M. Gay, and B.~W. Kernighan.
\newblock {\em {AMPL} : a modeling language for mathematical programming}.
\newblock Scientific Press, South San Francisco, 1993.

\bibitem{freuder}
E.~Freuder.
\newblock In pursuit of the holy grail.
\newblock {\em Constraints}, 2(1):57--61, 1997.

\bibitem{freuder2006constraint}
E.~C. Freuder and A.~K. Mackworth.
\newblock Constraint satisfaction: An emerging paradigm.
\newblock {\em Handbook of Constraint Programming}, pages 13--28, 2006.

\bibitem{Friesen:2015aa}
A.~Friesen and P.~Domingos.
\newblock Recursive decomposition for nonconvex optimization.
\newblock In {\em IJCAI}, 2015.

\bibitem{friesen2016sum}
A.~Friesen and P.~Domingos.
\newblock The sum-product theorem: A foundation for learning tractable models.
\newblock In {\em International Conference on Machine Learning}, pages
  1909--1918, 2016.

\bibitem{DBLP:journals/constraints/FrischHJHM08}
A.~M. Frisch, W.~Harvey, C.~Jefferson, B.~M. Hern{\'{a}}ndez, and I.~Miguel.
\newblock Essence : {A} constraint language for specifying combinatorial
  problems.
\newblock {\em Constraints}, 13(3):268--306, 2008.

\bibitem{modelcountingchapter}
C.~P. Gomes, A.~Sabharwal, and B.~Selman.
\newblock Model counting.
\newblock In {\em Handbook of Satisfiability}. IOS Press, 2009.

\bibitem{DBLP:journals/coling/Goodman99}
J.~Goodman.
\newblock Semiring parsing.
\newblock {\em Computational Linguistics}, 25(4):573--605, 1999.

\bibitem{conf/uai/GoodmanMRBT08}
N.~D. Goodman, V.~K. Mansinghka, D.~M. Roy, K.~Bonawitz, and J.~B. Tenenbaum.
\newblock Church: a language for generative models.
\newblock In {\em Proc. UAI}, pages 220--229, 2008.

\bibitem{gb08}
M.~Grant and S.~Boyd.
\newblock Graph implementations for nonsmooth convex programs.
\newblock In {\em Recent Advances in Learning and Control}, Lecture Notes in
  Control and Information Sciences, pages 95--110. Springer-Verlag Limited,
  2008.

\bibitem{green2007provenance}
T.~J. Green, G.~Karvounarakis, and V.~Tannen.
\newblock Provenance semirings.
\newblock In {\em Proceedings of the twenty-sixth ACM SIGMOD-SIGACT-SIGART
  symposium on Principles of database systems}, pages 31--40. ACM, 2007.

\bibitem{halmos-measure}
P.~Halmos.
\newblock {Measure theory}.
\newblock {\em Van Nostrad Reinhold Company}, 1950.

\bibitem{halpern1990analysis}
J.~Halpern.
\newblock {An analysis of first-order logics of probability}.
\newblock {\em Artificial Intelligence}, 46(3):311--350, 1990.

\bibitem{1384411}
H.~Hindi.
\newblock A tutorial on convex optimization.
\newblock In {\em American Control Conference}, volume~4, June 2004.

\bibitem{DBLP:journals/dam/HookerO99}
J.~N. Hooker and M.~A.~O. Lama.
\newblock Mixed logical-linear programming.
\newblock {\em Discrete Applied Mathematics}, 96-97:395--442, 1999.

\bibitem{kannan1978computational}
R.~Kannan and C.~L. Monma.
\newblock {\em On the computational complexity of integer programming
  problems}.
\newblock Springer, 1978.

\bibitem{kautzunified}
H.~Kautz.
\newblock Toward a universal inference engine.
\newblock In {\em LPNMR}, volume 2923 of {\em LNCS}, pages 2--2. Springer
  Berlin Heidelberg, 2004.

\bibitem{DBLP:conf/aaai/KimmigBR11}
A.~Kimmig, G.~Van~den Broeck, and L.~De~Raedt.
\newblock An algebraic prolog for reasoning about possible worlds.
\newblock In {\em Proc. AAAI}, 2011.

\bibitem{AMC}
A.~Kimmig, G.~Van~den Broeck, and L.~De~Raedt.
\newblock Algebraic model counting.
\newblock {\em Journal of Applied Logic}, 22:46--62, 2017.

\bibitem{books/daglib/0023091}
D.~Koller and N.~Friedman.
\newblock {\em Probabilistic Graphical Models - Principles and Techniques.}
\newblock MIT Press, 2009.

\bibitem{DBLP:conf/ifip/Kowalski74}
R.~A. Kowalski.
\newblock Predicate logic as programming language.
\newblock In {\em {IFIP} Congress}, pages 569--574, 1974.

\bibitem{MR98m:68152}
W.~Kuich.
\newblock Semirings and formal power series: their relevance to formal
  languages and automata.
\newblock In {\em Handbook of formal languages, Vol.\ 1}, pages 609--677.
  Springer, Berlin, 1997.

\bibitem{DBLP:conf/aaai/LierlerT15}
Y.~Lierler and M.~Truszczynski.
\newblock An abstract view on modularity in knowledge representation.
\newblock In {\em AAAI}, pages 1532--1538, 2015.

\bibitem{DBLP:conf/kr/LiuJN12}
G.~Liu, T.~Janhunen, and I.~Niemel{\"{a}}.
\newblock Answer set programming via mixed integer programming.
\newblock In {\em KR}, 2012.

\bibitem{Marriott:2008aa}
K.~Marriott, N.~Nethercote, R.~Rafeh, P.~Stuckey, M.~Garcia de~la Banda, and
  M.~Wallace.
\newblock The design of the zinc modelling language.
\newblock {\em Constraints}, 13(3):229--267, 2008.

\bibitem{DBLP:conf/ijcai/MilchMRSOK05}
B.~Milch, B.~Marthi, S.~J. Russell, D.~Sontag, D.~L. Ong, and A.~Kolobov.
\newblock {BLOG}: Probabilistic models with unknown objects.
\newblock In {\em Proc. IJCAI}, pages 1352--1359, 2005.

\bibitem{DBLP:conf/aaai/MitchellT05}
D.~G. Mitchell and E.~Ternovska.
\newblock A framework for representing and solving {NP} search problems.
\newblock In {\em AAAI}, pages 430--435, 2005.

\bibitem{nieuwenhuis2006sat}
R.~Nieuwenhuis and A.~Oliveras.
\newblock On sat modulo theories and optimization problems.
\newblock In {\em Theory and Applications of Satisfiability Testing-SAT 2006},
  pages 156--169. Springer, 2006.

\bibitem{Orsini2017}
F.~Orsini, P.~Frasconi, and L.~De~Raedt.
\newblock kproblog: an algebraic prolog for machine learning.
\newblock {\em Machine Learning}, 106(12):1933--1969, Dec 2017.

\bibitem{poon2011sum}
H.~Poon and P.~Domingos.
\newblock Sum-product networks: A new deep architecture.
\newblock In {\em 2011 IEEE International Conference on Computer Vision
  Workshops (ICCV Workshops)}, pages 689--690. IEEE, 2011.

\bibitem{richardson2006markov}
M.~Richardson and P.~Domingos.
\newblock Markov logic networks.
\newblock {\em Machine learning}, 62(1):107--136, 2006.

\bibitem{sankaranarayanan2013static}
S.~Sankaranarayanan, A.~Chakarov, and S.~Gulwani.
\newblock Static analysis for probabilistic programs: inferring whole program
  properties from finitely many paths.
\newblock {\em ACM SIGPLAN Notices}, 48(6), 2013.

\bibitem{DBLP:conf/ijcai/SannerM05}
S.~Sanner and D.~A. McAllester.
\newblock Affine algebraic decision diagrams (aadds) and their application to
  structured probabilistic inference.
\newblock In {\em IJCAI}, pages 1384--1390, 2005.

\bibitem{Sebastiani:2015:OMT:2737801.2699915}
R.~Sebastiani and S.~Tomasi.
\newblock Optimization modulo theories with linear rational costs.
\newblock {\em ACM Trans. Comput. Logic}, 16(2):12:1--12:43, Feb. 2015.

\bibitem{DBLP:conf/icra/SrivastavaFRCRA14}
S.~Srivastava, E.~Fang, L.~Riano, R.~Chitnis, S.~J. Russell, and P.~Abbeel.
\newblock Combined task and motion planning through an extensible
  planner-independent interface layer.
\newblock In {\em ICRA}, pages 639--646, 2014.

\bibitem{DBLP:conf/ijcai/TernovskaM09}
E.~Ternovska and D.~G. Mitchell.
\newblock Declarative programming of search problems with built-in arithmetic.
\newblock In {\em Proc. IJCAI}, pages 942--947, 2009.

\bibitem{thrun2005probabilistic}
S.~Thrun, W.~Burgard, and D.~Fox.
\newblock {\em Probabilistic Robotics}.
\newblock {MIT Press}, 2005.

\bibitem{DBLP:journals/informs/Hentenryck02}
P.~Van~Hentenryck.
\newblock Constraint and integer programming in {OPL}.
\newblock {\em {INFORMS} Journal on Computing}, 14(4):345--372, 2002.

\end{thebibliography}

\end{document}